\documentclass[10pt,twocolumn,letterpaper]{article}

\usepackage[pagenumbers]{cvpr} 

\definecolor{cvprblue}{rgb}{0.21,0.49,0.74}
\usepackage[pagebackref,breaklinks,colorlinks,allcolors=cvprblue]{hyperref}
\usepackage{amsmath}
\usepackage{algorithm}
\usepackage{algorithmic}
\usepackage{booktabs}
\usepackage[table]{xcolor}
\usepackage{tabularray}
\usepackage{pifont}

\def\paperID{986} 
\def\confName{CVPR}
\def\confYear{2026}

\title{Disc3D: Automatic Curation of High-Quality 3D Dialog Data via Discriminative Object Referring}

\author{
Siyuan Wei\textsuperscript{1} \quad
Chunjie Wang\textsuperscript{1} \quad
Xiaosheng Yan\textsuperscript{1} \quad
Rui Huang\textsuperscript{2} \quad
Zhishan Zhou\textsuperscript{1} \quad
Xiao Liu\textsuperscript{1} \\
\textsuperscript{1}PICO, ByteDance, Beijing \\
\textsuperscript{2}Tsinghua University \\
{\tt\small \{weisiyuan.buaa, wangchunjie01, yanxiaosheng, zhouzhishan.2013,liuxiao.ai\}@bytedance.com} \\
{\tt\small hr20@mails.tsinghua.edu.cn}
}

\begin{document}
\maketitle

\begin{abstract}
3D Multi-modal Large Language Models (MLLMs) still lag behind their 2D peers, largely because large-scale, high-quality 3D scene–dialogue datasets remain scarce. Prior efforts hinge on expensive human annotation and leave two key ambiguities unresolved: viewpoint ambiguity---spatial language presumes unknown camera poses, and object referring ambiguity---non-exclusive descriptions blur the line between targets and distractors. We therefore present a fully automated pipeline that converts raw 3D scans into unambiguous, high-quality dialogue data at a fraction of the previous cost. By synergizing rule-based constraints with 2D MLLMs and LLMs, the pipeline enables controllable, scalable generation without human intervention. The pipeline comprises four stages: (1) meta-annotation collection harvesting object-, frame-, and scene-level captions, (2) scene graph construction with relation correction to capture the proximal object relations, (3) discriminative object referring that generates exclusive and compact descriptions, and (4) multi-task data generation synthesizing diverse dialogues. Our pipeline systematically mitigates inherent flaws in source datasets and produces the final Disc3D dataset---over 2 million samples in 25K hybrid 3D scenes, spanning scene\&view\&object captioning, visual grounding, and five object-centric QA tasks. Extensive experiments demonstrate that training with Disc3D yields consistent, significant improvements on both public benchmarks and our multifaceted Disc3D-QA tasks. Our codebase is available at \url{https://github.com/bytedance/Disc3D},
and the dataset at \url{https://huggingface.co/datasets/Sywwwwww/Disc3D}.

\end{abstract}    

\section{Introduction}
\label{sec:intro}

Multi-modal understanding of 3D scenes is a prerequisite for embodied agents and robotic systems, yet current models lag far behind their 2D counterparts.
A key bottleneck is the lack of versatile 3D vision foundation models, exacerbated by the acute scarcity of large-scale, annotated 3D scene dialogues. Compared with 2D data, 3D modality incurs prohibitive collection and annotation costs, severely limiting the scale of datasets that can support training 3D Multi-modal Large Language Models (MLLMs).

\begin{figure}[t]
\centering
\includegraphics[width=0.45\textwidth]{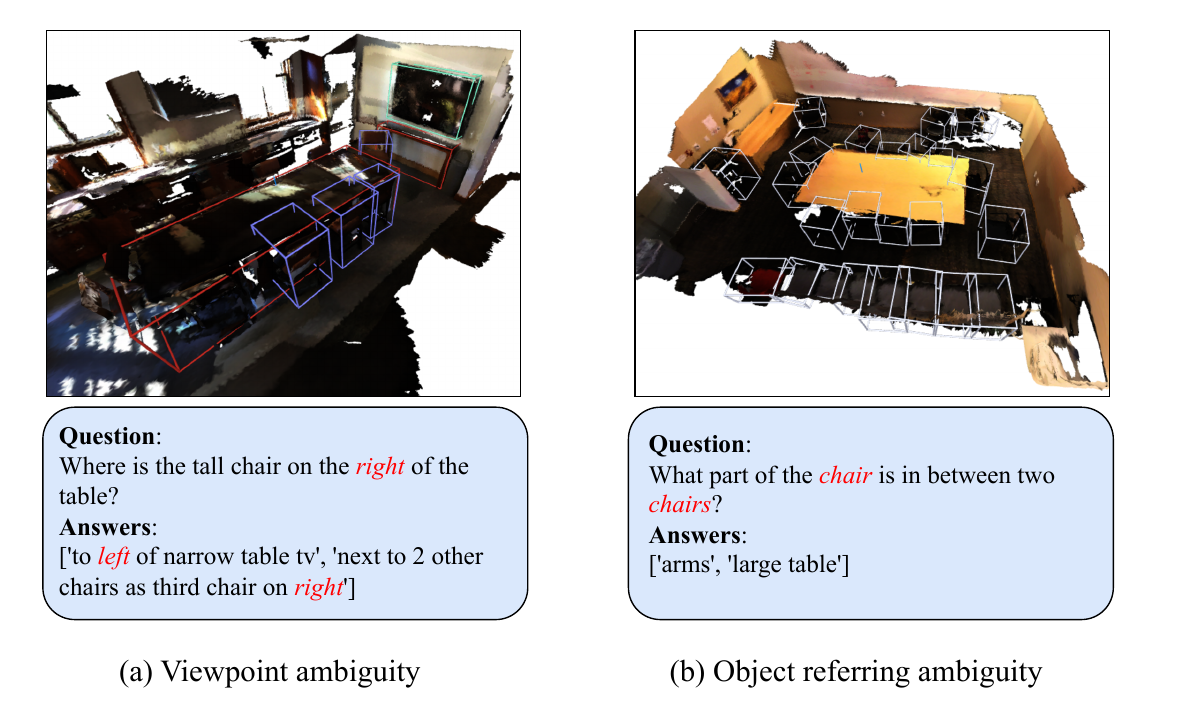} 
\caption{ Examples of ambiguity in ScanQA, including (a) \emph{viewpoint ambiguity}: the relative positions of objects (e.g., right or left) implied in a dialogue depend on viewpoint information not present in the context; and (b) \emph{object referring ambiguity}: object descriptions (e.g., chair) lack discriminative detail, resulting in confusion between the target object and distractors.}
\label{fig/ds_drawback}
\end{figure}

\begin{figure}[t]
\centering
\includegraphics[width=0.45\textwidth]{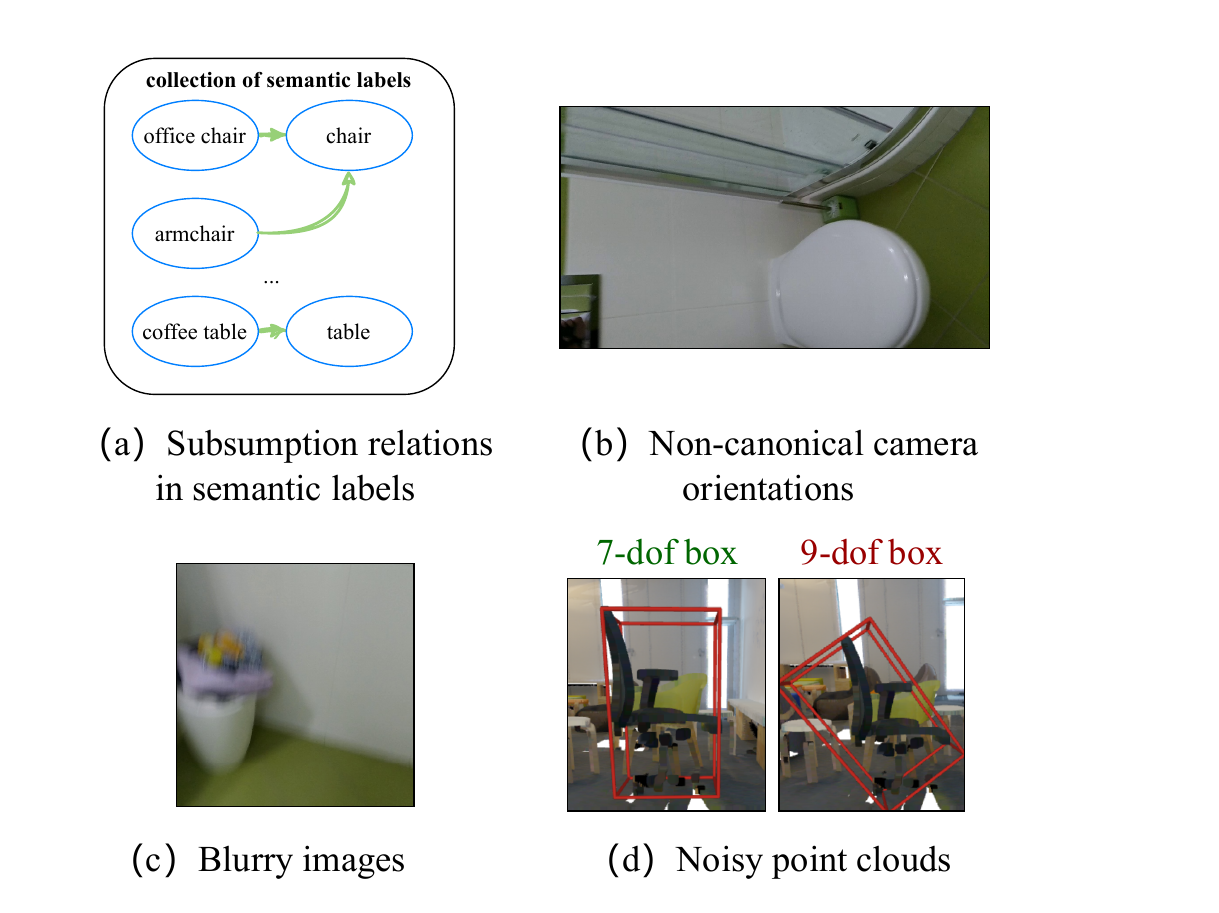} 
\caption{Common defects in 3D scene datasets that impede dialogue data generation: a) \emph{Semantic label subsumption} (e.g., \textit{office chair} vs. \textit{chair}) induces referential ambiguity; b) \emph{Non-canonical camera views} degrade 2D MLLM annotation accuracy; c) \emph{Blurry images} reduce visual confidence; d) \emph{Noisy point clouds} produce inaccurate 9-DOF boxes, impairing downstream learning.}
\label{fig/src_def}
\end{figure}

Existing 3D MLLM datasets predominantly rely on manual annotation~\cite{lyu2024mmscan,azuma2022scanqa,ma2022sqa3d,achlioptas2020referit3d,chen2020scanrefer,zhang2023multi3drefer,yang2025thinking}, which is inherently costly and limits scalability. Moreover, free-form annotations introduce two critical ambiguities: 1)\emph{Viewpoint ambiguity}: As \cref{fig/ds_drawback}a shows, spatial descriptions often presuppose a specific 2D camera pose that is unknown during training and evaluation; 2) \emph{Object referring ambiguity}: The unconstrained object referring expressions lack discriminative features, making target objects indistinguishable from distractors, as shown in \cref{fig/ds_drawback}b. Meanwhile, as \cref{fig/src_def} shows, inherent deficiencies in existing 3D scene datasets substantially increase the difficulty of multi-modal data generation. These issues necessitate an automated, systematic, scalable pipeline to convert raw scans into unambiguous, high-quality 3D dialogue data.

\begin{figure*}[t]
\centering
\includegraphics[width=0.96\textwidth]{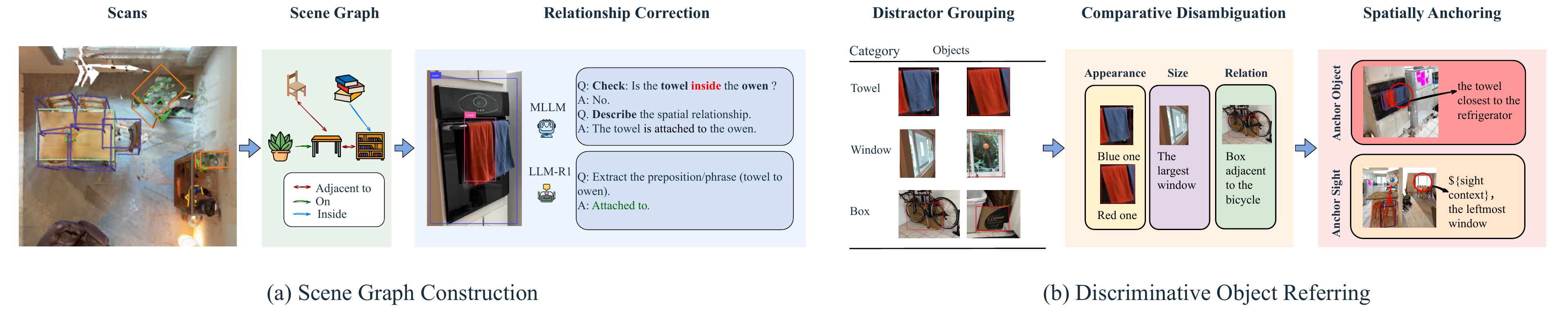}
\caption{A schematic overview of two key stages in our data curation pipeline. Specifically: a) Scene Graph Construction proceeds in two stages: an initial graph is built automatically, after which a LLM \& MLLM-guided module refines the misjudged relations (highlighted in red).
b) Discriminative Object Referring produces exclusive, unambiguous descriptions for each object in the distractor group across five orthogonal axes.  
The first sub-stage, Comparative Disambiguation, contrasts objects of the same category along appearance, size, and relational cues; the second sub-stage, Spatially Anchoring, injects 3D context by explicitly conditioning every description on the designated anchor object or sight.}

\label{fig:ppl}
\end{figure*}

To tackle these challenges, we present the \emph{Disc3D} dataset, which features a fully automated pipeline for generating high-quality 3D scene dialog data. The pipeline integrates 2D MLLM and LLM capabilities with rule-based processes at each stage, thereby leveraging the model’s capacity to perform automatic annotation under a controlled paradigm. Drawing inspiration from how humans identify an object's distinctive features to resolve ambiguity in real-world conversations, we introduce \textit{Discriminative Object Referring}. It enhances the dialogue's contextual information by focusing on object referrals, reduces potential referring ambiguities, and increases the task's complexity. Specifically, our pipeline consists of four stages:
\begin{enumerate}
  \item \textbf{Meta-Annotation Collection}: Acquire scene-, image-, and object-level annotations by prompting 2D MLLMs to support subsequent multi-task generation.
  \item \textbf{Scene Graph Construction}: Construct a spatial relationship graph (objects as nodes) using rule-based methods, then refine it to correct errors arising from rule misjudgments as shown in \cref{fig:ppl}a.
  \item \textbf{Discriminative Object Referring}: Generate exclusive descriptions for each object along five dimensions to distinguish objects from possible distractors as shown in \cref{fig:ppl}b.
  \item \textbf{Multi-Task Data Generation}: Leverage the above information to produce diverse, context-rich dialogue data.
\end{enumerate}

\begin{figure*}[t]
\centering
\includegraphics[width=0.96\textwidth]{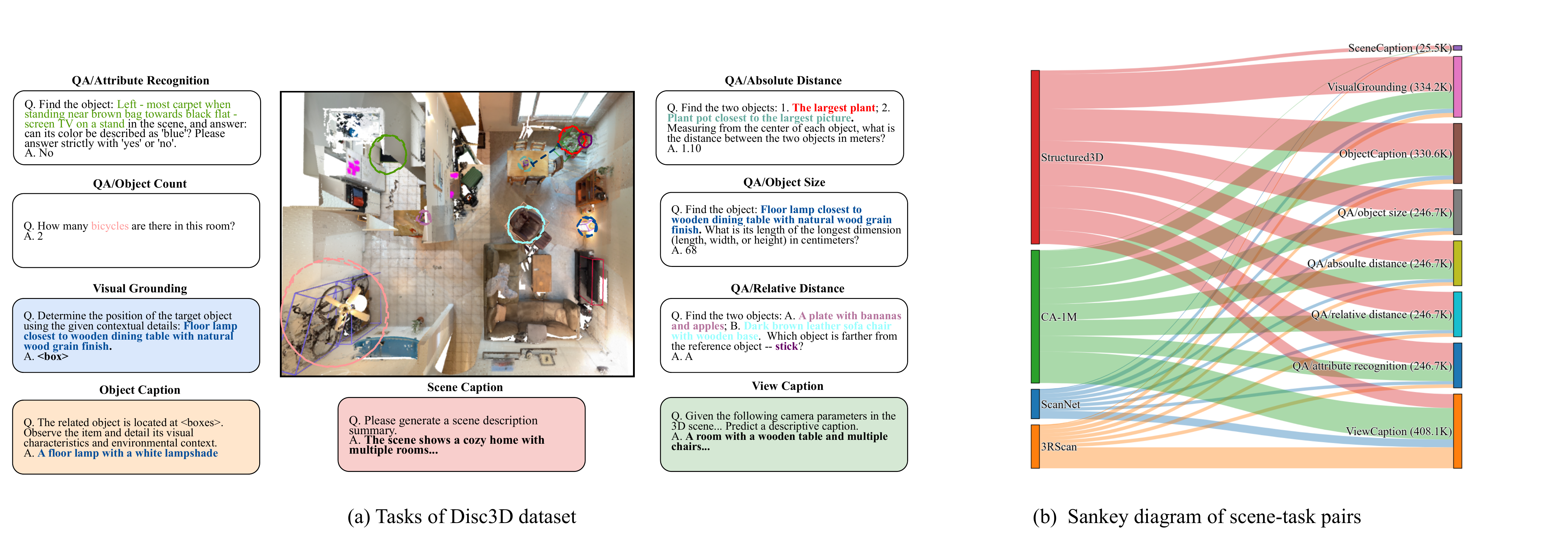}
\caption{Overview of our Disc3D dataset. 
(a) Disc3D comprises millions of dialogue samples across 25K hybrid (real and synthetic) scenes, covering five object-centric QA tasks and three caption tasks for training \& benchmarking 3D MLLMs. 
Object descriptions are color-coded to match the corresponding highlighted objects. 
(b) Task and scene distributions in the training set. The object counting task is excluded due to category-specific answer bias.}

\label{fig:Disc3D}
\end{figure*}

During production, we specifically address inherent issues in source datasets as shown in \cref{fig/src_def}. Noting the absence of a fully open-source 3D MLLM data curation pipeline, we will release our entire pipeline to foster community development. Utilizing this pipeline, we construct \emph{Disc3D} dataset, as shown in \cref{fig:Disc3D}, comprising millions of multi-task dialogue samples in 25K scans. We prove that training established 3D MLLM architectures on Disc3D yields promising improvements on our benchmark and other public benchmarks.

Our contributions are summarized as follows:
\begin{itemize}
    \item We introduce a fully automated 3D scene dialog data curation pipeline. Centered on discriminative object referring, it generates context-rich, high-quality multi-task data while minimizing the impact of source dataset flaws and avoiding possible ambiguities.
    \item We present \emph{Disc3D}, a million-scale dataset for training \& evaluating 3D MLLMs, encompassing diverse and comprehensive range of task types.
    \item We re-implement one of state-of-the-art 3D MLLMs and utilize the Disc3D dataset for training. Extensive evaluations on Disc3D and other existing benchmarks show that our Disc3D dataset leads to substantial performance improvements.
\end{itemize}

\section{Related Work}
\label{sec:related_work}

\paragraph{3D Multi-modal Datasets.}
Early work exploited 3D scene scans~\cite{dai2017scannet,wald2019rio,chang2017matterport3d,procthor,baruch2021arkitscenes,ramakrishnan2021hm3d,mao2022multiscan} to establish diverse 3D multi-modal benchmarks, including 3D question answering (QA)~\cite{azuma2022scanqa,ma2022sqa3d,yang2025thinking,OpenEQA2023} and 3D visual grounding~\cite{chen2020scanrefer,achlioptas2020referit3d,zhang2023multi3drefer}. 3D QA task assesses 3D MLLM’s understanding of objects and layouts in the scene~\cite{zhu20233d,chen2024ll3da,huang2024chat}, while 3D visual grounding tasks evaluate its ability to localize objects ~\cite{yuan2021instancerefer,yang2021sat,huang2021text,luo20223d,wu2023eda,xu2024vlm,chen2023unit3d}. Motivated by advances in Large Language Models (LLMs) and 2D Multi-modal LLMs (MLLMs), more recent efforts~\cite{jia2024sceneverse,lyu2024mmscan} combine human annotations with 2D MLLM capabilities to generate large-scale 3D scene data. However, these pipelines suffer from viewpoint ambiguity and referential uncertainty due to unconstrained annotations, and incur high production costs because of extensive manual labeling. We address these challenges with a fully automated pipeline based on Discriminative Object Referring, producing a larger, more diverse corpus free from previous ambiguities and cost constraints.

\paragraph{3D Multi-modal Large Language Models.}
Building on the rapid progress of 2D MLLMs~\cite{bai2025qwen2,zhu2024llava,wu2024deepseek,team2025Kimi}, recent research integrates 3D visual information into LLMs to create 3D MLLMs. Proposed methods include point‑cloud modules~\cite{avetisyan2024scenescript,deng20253d,xu2024pointllm,qi2024gpt4point}, lifted 2D foundation features~\cite{zhu2024llava}, video‑based representations~\cite{zheng2025video}, instance‑level embeddings~\cite{yu2025inst3d}, and bird’s‑eye‑view projections~\cite{GPT4Scene}. Owing to the increased dimensionality and complexity of 3D data, no single architecture yet demonstrates robust spatial understanding from limited training samples. 

\section{ Dataset }
\label{sec/data_curation_ppl}
This section first reflects on key issues in prior data work within the 3D MLLM domain. Next, we discuss the inherent flaws of existing 3D scene datasets that constrain data generation. We then describe our automated curation pipeline and conclude with a statistical profile and analysis of the resulting Disc3D corpus.

\subsection{Reflections on Existing 3D MLLM Datasets}

\label{subsec/reflection}
During our study of 3D MLLMs, we observed two major deficiencies in existing datasets as illustrated in \cref{fig/ds_drawback}:

\begin{itemize}
    \item \textit{Viewpoint Ambiguity:} Relative positional descriptions within the context (e.g., front/back, left/right) are tied to the camera viewpoint observed by the annotator or MLLM. While conventional references (e.g., adjacent walls) sometimes exist, the absence of explicit viewpoint information hinders the model training or evaluation in reasoning the spatial relationship.

    \item \textit{Object Referring Ambiguity:} The unconstrained description (e.g., the object category or caption), when used for object referring, lacks exclusivity and may be confused with other objects in the scene.
\end{itemize}

\subsection{Deficiencies in Source Data}

Existing 3D scene datasets were not created with 3D MLLMs in mind, and their idiosyncrasies must be explicitly neutralised before they can be reused for multi-modal instruction tuning (\cref{fig/src_def}).  
Two obstacles are particularly problematic:

\begin{itemize}
  \item \emph{Subsumption relations in semantic labels.}  
  Class names often form a taxonomy: \emph{dining table} is a hyponym of \emph{table}.  
  Treating every hypernym as a stand-alone label either forces the model to overlook valid hyponyms or invites context-level ambiguity when only the broader term is accepted as correct.

  \item \emph{Non-canonical camera poses.} Arbitrary viewpoints--especially those with large out-of-plane rotation---systematically degrade the quality of MLLM-generated annotations.
\end{itemize}

Because the label vocabularies differ across source corpora, we let a strong LLM~\cite{liu2024deepseek} build a single directed graph that encodes the subsumption structure.  
Each node is an object label; a directed edge $u \rightarrow v$ exists whenever LLM confirms that $v$ is a subtype of $u$ (\cref{fig/src_def}a).  
Candidate edges are first shortlisted by word overlap and then validated by LLM in a zero-shot fashion.

For non-canonical orientations, we rotate images in $90^\circ$ increments to approximate a canonical horizontal viewpoint. Other issues, such as point cloud noise and image blur, are mitigated via re-calculating 7-DOF bounding boxes of annotated objects and tailored prompt engineering for 2D MLLMs. Further implementation details are provided in the supplementary material.

\begin{algorithm}[tb]
\caption{Comparative Disambiguation}
\label{alg:cdor_v2}
\textbf{Input}: The distractor group $O = \{o_1, o_2, \dots, o_n\}$.\\
\textbf{Output}: A mapping $M$ from each object $o_i$ to its descriptions.
\begin{algorithmic}[1]
\STATE \textit{\# Step 1. Group objects via multi-aspect comparison}
\STATE \textit{\# $G_{\text{app}}$: appearance (e.g., 'red') $\mapsto$ objects }
\STATE $G_{\text{app}} \gets \text{GroupByAppearance}(O, \text{LLM})$ 
\STATE \textit{\# $G_{\text{size}}$: size (e.g., 'the largest') $\mapsto$ objects }
\STATE $G_{\text{size}} \gets \text{GroupBySize}(O)$ 
\STATE \textit{\# $G_{\text{rel}}$:  object relations (e.g., 'beside the table') $\mapsto$ objects }
\STATE $G_{\text{rel}} \gets \text{GroupByRelations}(O)$ 
\STATE $G_{\text{all}} \gets$ merge($G_{\text{app}}, G_{\text{size}}, G_{\text{rel}}$)

\STATE \textit{\# Step 2. Assign exclusive and compact descriptions}
\STATE $C \gets \{ k \mid k \in G_{\text{all}}[\text{k}] \}$
\FOR{each object $o_i \in O$}
    \STATE Initialize $D_i = \emptyset$ for object $o_i$
    \FOR{each non-empty subset $S \subseteq C$}
        \STATE $U \gets \bigcap_{k \in S} G_{\text{all}}[k]$
        \IF{$U = \{o_i\}$}
            \STATE $d_S \gets \text{", ".join}(\text{sorted}(S))$
            \STATE Append $d_S$ to $D_i$
        \ENDIF
    \ENDFOR
    \STATE \textit{\# Keep only compact descriptions}
    \STATE \textit{\# (e.g., prefer "red" over "red, large").}
    \STATE $M[o_i] \gets \{\, d \in D_i \mid \text{is\_compact}(d) \,\}$
\ENDFOR

\STATE \textbf{return} $M$
\end{algorithmic}
\end{algorithm}

\subsection{Data Curation Pipeline}
\label{subsection:data_curation_ppl}

For scene acquisition, we compile 25\,K scans from five datasets: ScanNet~\cite{dai2017scannet}, CA-1M~\cite{lazarow2025cubify}, 3RScan~\cite{wald2019rio}, Structured3D~\cite{zheng2020structured3d} (training), and ScanNet++~\cite{yeshwanthliu2023scannetpp} (testing), covering both real and synthetic environments.  
The subsequent curation pipeline consists of four stages.

\paragraph{Meta-annotation collection.} Generally, we prompt our in-house MLLMs and LLMs to produce object-, frame-, and scene-level captions during this step. The specific steps are as follows.
\begin{itemize}
    \item \textit{Frame sampling}. To reduce latency and token cost, we subsample each scan.  
  Video sequences are uniformly sampled; the frames from non-video sequences (e.g. 3RScan) are clustered by camera pose, and one representative frame is retained per cluster.

    \item \textit{Caption annotation.}  For scenes \& frames, we prompt MLLM with given images and instructions, while max-coverage sampling~\cite{zheng2025video} caps the number of frames used for the scene caption. For objects, we rank viewpoints by (i) the 2D distance from the object center to the image center and (ii) the visible area ratio obtained by rendering the annotated mesh. The top-2 views are cropped and sent to the MLLM for captioning; an LLM then extracts appearance attributes. 
\end{itemize}

\paragraph{Scene Graph Construction.}
Following SceneVerse~\cite{jia2024sceneverse}, we approximate object occupancy with 3D bounding boxes and build an object-centric graph that records only vertical adjacency and horizontal contact.
To suppress noise and ambiguity, we discard multi-object relations (involving more than two nodes) and all relative directions (e.g., left/right or clock-face positions).

Rule-based heuristics cannot identify subtle cases such as ``inside'' or ``hanging from''.
We therefore introduce a \emph{Relation Correction} stage, illustrated in \cref{fig:ppl}a.
An MLLM first narrates the interaction between co-visible objects in free text; an R1-style LLM~\cite{guo2025deepseek} then parses each sentence into one of the predefined relation templates, updating the initial graph accordingly.

\begin{figure}[t]
\centering
\includegraphics[width=0.45\textwidth]{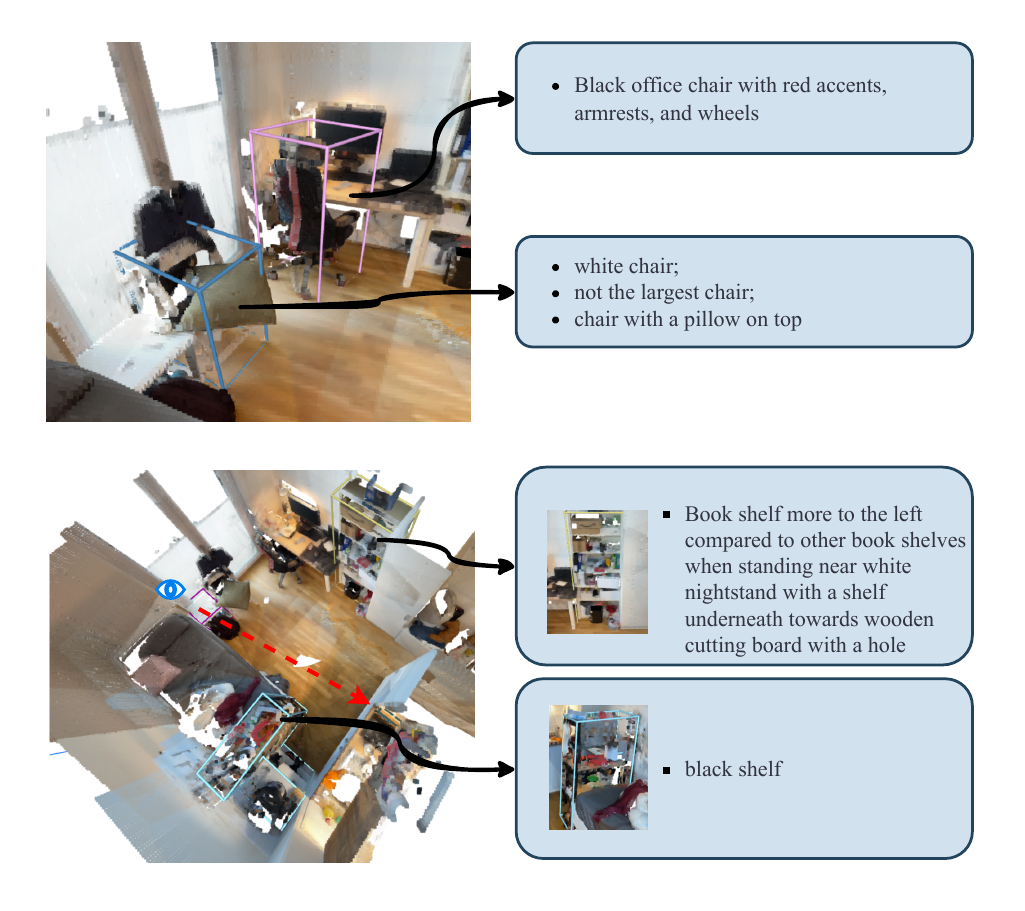}
\caption{Discriminative object referral examples in Scannet++~\cite{yeshwanthliu2023scannetpp} scans.}
\label{fig:obj_ref_cases}
\end{figure}

\paragraph{Discriminative Object Referring.}
To mimic the referential adjustments that emerge naturally in human conversation, we exploit the object-level annotations and scene graphs acquired earlier to produce concise, distinguishing descriptions that separate each target from its distractors (\cref{fig:ppl}b). The proposed method consists of the following three steps to assign non-conflicting referring expressions to similar objects (see cases in ~\cref{fig:obj_ref_cases}).

\begin{itemize}
    \item \textit{Distractor Grouping}: Objects are clustered by categorical relatedness.  
    Using the label graph as prior, an object is added to a group when its label is identical to, or a hyponym of, the group label. This step isolates the set of objects that can plausibly be confused with one another.
    
    \item \textit{Comparative disambiguation.}  
    For every distractor group, we derive a minimal set of modifiers as the initial discriminative object referrals (see the pseudocode in \cref{alg:cdor_v2}).
    \begin{description}
        \item[1] \textit{Multi-aspect comparison}: For the appearance dimension, LLM is prompted with few-shot examples to cluster objects by attribute differences. We provide few-shot labeling examples in the prompt to guide the model's annotation behavior and verify that LLM's formatted output meets the requirements. For the size dimension, 7-DOF box volumes are compared; a generous tolerance compensates for box–object mismatch. For the object relation dimension, object-centric sub-scene graphs are contrasted to extract distinctive relations. 
        \item[2] \textit{Descriptor fusion}: We combine the results from these dimensions to obtain \( G_{\text{all}} \), which represents the subset of objects that match the descriptors. Given \( G_{\text{all}} \), we can derive exclusive and compact descriptive modifiers for each object, distinguishing it from others without any redundant descriptors (see Algorithm~\ref{alg:cdor_v2}, lines~9--END).
        \item[3] \textit{Singleton fallback}: For objects without distractors, we use their captions or labels directly. The resulting descriptors are rewritten by LLM to reduce template bias. 
    \end{description}
    
    \item \textit{Spatially anchoring}: To incorporate richer 3D context, we introduce two anchors: the \emph{anchor object} and \emph{anchor sight}. For each distractor group, we identify the nearest (or farthest) object relative to the anchor object, and the leftmost (or rightmost) object relative to the anchor sight connecting two objects. All objects serving as references are sampled from non-distractor objects that have already been assigned discriminative object referrals by Comparative disambiguation.
    Spatial clauses are first templated, then rewritten by LLM for fluency; supplementary material gives full implementation details.
\end{itemize}

\paragraph{Multi-Task Data Generation.} As \cref{fig:Disc3D} shows, our data curation pipeline yields three caption tasks: \textit{scene caption}, \textit{object caption}, and \textit{view caption}. The \emph{view caption} task requires 3D MLLMs to summarize the visible content with the given camera pose and intrinsics. Discriminative object referrals feed the \textit{visual grounding}. 

For QA, we define five subtasks: \textit{object size}, \textit{absolute distance}, \textit{relative distance}, \textit{object count}, and \textit{attribute recognition} to cover the all-around spatial understanding capabilities. All questions are instantiated from templates using discriminative object referrals. For the \emph{attribute recognition} task, we restrict the object referral to either the semantic label of the queried object (when no distractors are present) or a spatially anchored object referral, thereby eliminating any risk of answer leakage. This QA task subsumes two formats: True-False items and open-ended queries. For the True-False item with the wrong attribute, a plausible yet incorrect property is generated by prompting LLM for the queried object. To compute the distance between two objects for relevant QA tasks, we sample a dense set of points from the surface of 7-DOF bounding boxes. The distance is then defined as the minimum Euclidean distance between these two point sets. Further details are provided in the supplementary material.

\begin{table*}[t]
  \centering
  \label{tab:dataset-comparison}
  \small
  \setlength{\tabcolsep}{1mm}
  \begin{tblr}{
      hline{1,5} = {-}{0.08em},
      hline{2} = {-}{},
      row{4} = {bg=green!5}
    }
    Dataset      & Samples & Scans & Viewpoint-ambiguity-free & Human-annotation-free & Object Referral Constraint & Task Diversity \\
    SceneVerse~\cite{jia2024sceneverse}    & ~ ~2.5M & 68K   & No & No                 & No                         &        \ding{72}        \\
    MMScan~\cite{lyu2024mmscan}       & 1.45M   & 5.2K  & No & No                 & No                         &       \ding{72} \ding{72} \ding{73}         \\
    Disc3D, ours & 2.08M   & 25K   & Yes & Yes                 & Discriminative             &  \ding{72} \ding{72} \ding{72}
    \end{tblr}
      \caption{Comparison with two other large-scale 3D MLLM multi-task datasets. Disc3D autonomously curates a two-million-sample multi-task corpus by means of Discriminative Object Referencing, a carefully engineered module that injects contextual constraints into object descriptions and eliminates ambiguities in object identification and spatial relations. Earlier 3D vision-language datasets~\cite{azuma2022scanqa,ma2022sqa3d,achlioptas2020referit3d, zhang2023multi3drefer}, being smaller and task-specific, still remain vulnerable to ambiguity; they are therefore excluded from direct comparison.
    }
\label{tab:ds_cmp}
\end{table*}

\subsection{Statistics \& Analysis}
\label{subsec:stats}

\paragraph{Overview.} As illustrated in \cref{fig:Disc3D}, the Disc3D dataset aggregates 25\,K scans drawn from simulated and real-world sources. With an automated curation pipeline, we generate multi-million 3D scene dialogues in a matter of days; this count is by no means the pipeline’s ceiling. We perform sampling to ensure a balanced distribution across different QA sub-tasks. To prevent dialogues of complex scenes from dominating the dataset, we impose limits on the number of QA instances generated per scene. 

\paragraph{Comparison with existing large-scale 3D multi-modal datasets.}
MMScan~\cite{lyu2024mmscan} and SceneVerse~\cite{jia2024sceneverse} assemble large-scale 3D multi-modal datasets with human-in-the-loop pipelines.
Disc3D eliminates manual effort by replacing annotators with a rule-based engine powered by MLLMs and LLMs.
The resulting corpus spans more scenes and samples than MMScan and subsumes the task diversity of SceneVerse.
Like MMScan, Disc3D provides scene- and region-level captioning, visual grounding, and multi-dimensional QA.
We extend this set with view-level captioning derived from individual frames.
Throughout data curation, we eliminated viewpoint ambiguities introduced by either manual design or LLM-generated text.
To resolve object-referring ambiguity, we propose Comparative Disambiguation and Spatial Anchoring, which embed discriminative constraints into object descriptions.
A brief comparison is provided in \cref{tab:ds_cmp}.

\begin{figure}[t]
\centering
\includegraphics[width=0.48\textwidth]{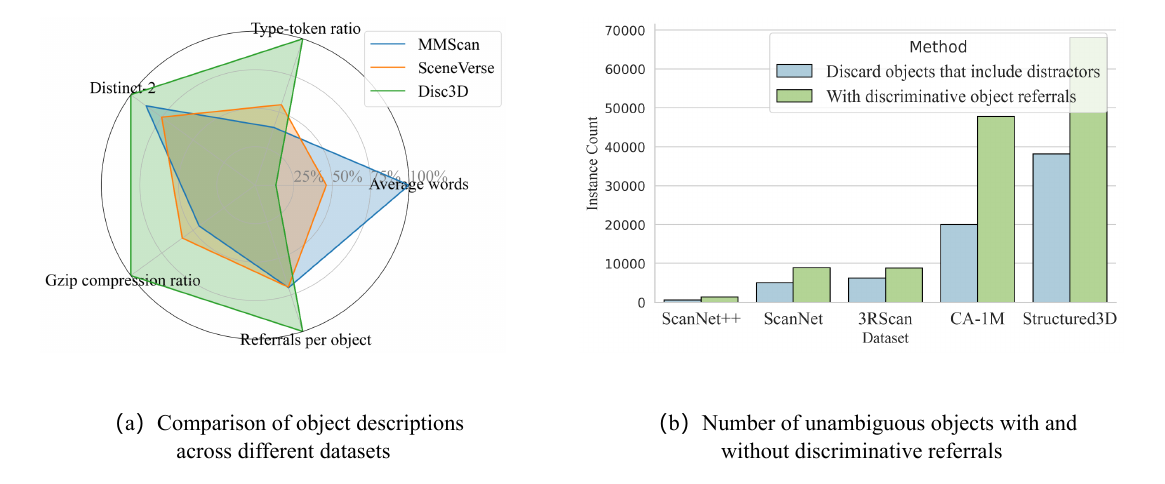}
\caption{Analysis of object referrals. (a) Our Disc3D incrementally enriches object references, incorporating textual modifiers chosen along carefully controlled dimensions, and thereby yields concise, information-rich object descriptions. The gzip compression ratio plotted is the ratio of the compressed size (bytes) to the original size. (b)~By explicitly modeling distractors, Discriminative Object Referring utilizes more object annotations in each source dataset in the absence of referential ambiguity.}
\label{fig:obj_ref_cmp}
\end{figure}

\paragraph{Analysis of Object Referral.}
When no referential ambiguity exists, a trivial baseline simply discards any object that has distractors, thereby avoiding confusion but wasting most annotations.  
In contrast, our \emph{Discriminative Object Referring} strategy lets the curation pipeline retain a substantially larger subset of annotated objects from the source dataset (\cref{fig:obj_ref_cmp}b).  
Moreover, compared with MMScan and SceneVerse, Disc3D generates for each object a richer yet concise collection of object descriptions (\cref{fig:obj_ref_cmp}a).

\paragraph{Human Revision for Test Split.} For the Disc3D test split, we conducted manual quality inspection and correction to evaluate the accuracy of object referrals. Since spatially anchored descriptions introduce minimal noise but exhibit low manual annotation accuracy, we focused on verifying object descriptions generated in the \emph{Comparative Disambiguation} stage. Specifically, annotators compared the 2D and 3D views of target and distractor objects to assess the factual accuracy and exclusiveness of the provided descriptions, revising any inaccuracies.

\begin{figure}[t]
\centering
\includegraphics[width=0.45\textwidth]{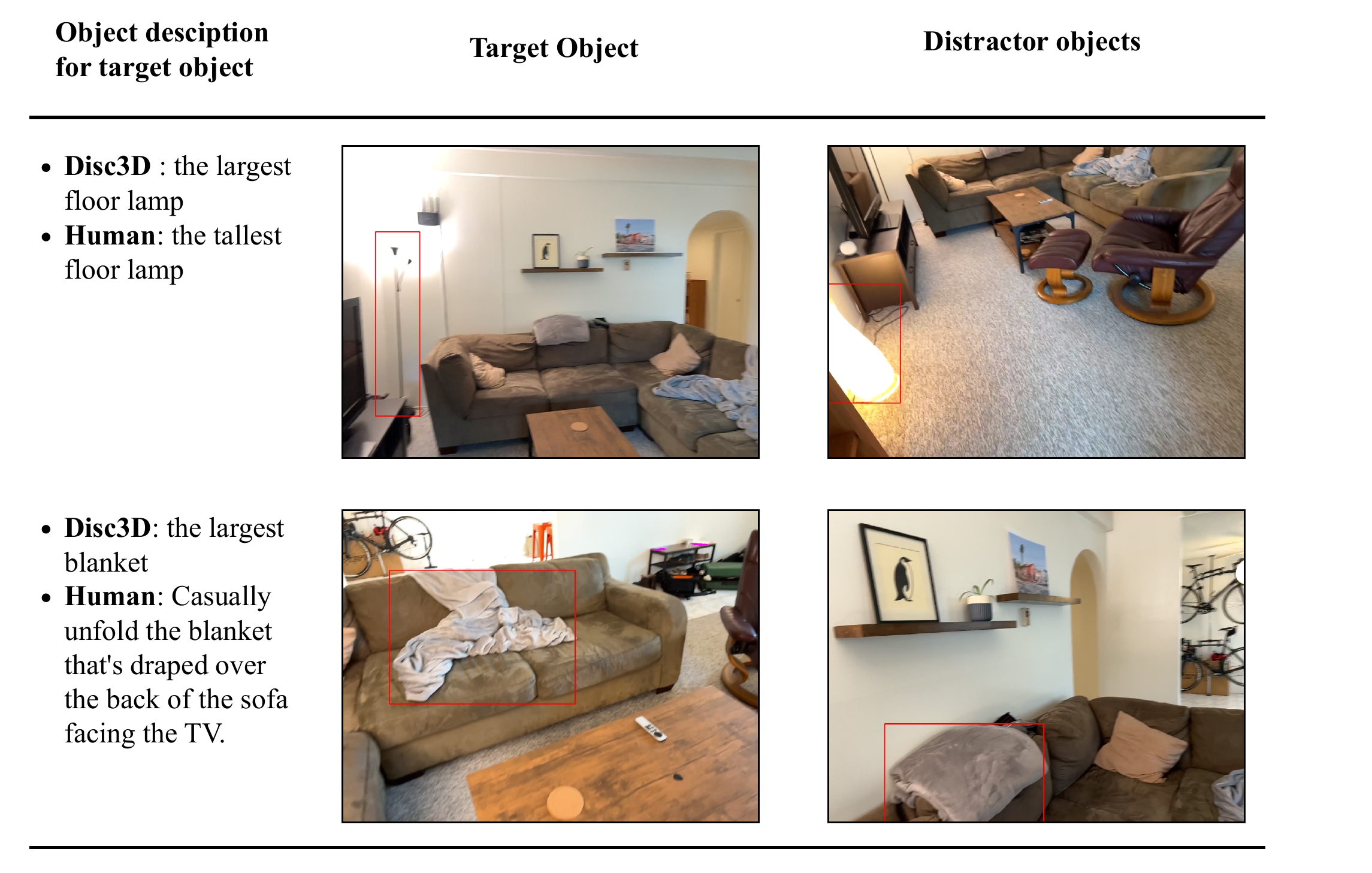}
\caption{Comparison of object descriptions before and after
manual modification.}
\label{fig:human_ann}
\end{figure}

In total, 12\% of the descriptions ($130/1092$) were revised.
As illustrated in \cref{fig:human_ann}, the majority of edits introduced by human annotators do not rectify factual inaccuracies but instead refine imprecise wording.
For instance, limited descriptors such as ``largest'' or ``smallest'' often inadequately capture the distinctive size characteristics of irregularly shaped objects (e.g., floor lamps) or soft, deformable items (e.g., blankets).
We contend that such imprecisions, though potentially influencing model preferences, are comparatively tolerable relative to factual errors or ambiguous expressions.
Given that manually annotated benchmarks still exhibit non-negligible noise---e.g., human re-evaluation in~\cite{huang2025unveiling} reports only 62\% and 80\% accuracy on ScanQA and SQA3D, respectively---Disc3D provides 3D multi-modal data with comparatively low label noise.

Finally, the manually corrected descriptions replaced the original ones for all subsequent processing, yielding the final benchmark of Disc3D.

\begin{table*}[ht]
\centering
\small
\setlength{\tabcolsep}{1mm}
\begin{tblr}{
  row{odd} = {c},
  row{2} = {c},
  row{4} = {c},
  row{6} = {c},
  row{8} = {c},
  cell{1}{1} = {r=2}{},
  cell{1}{2} = {r=2}{},
  cell{1}{3} = {c=5}{},
  cell{1}{8} = {r=2}{},
  cell{1}{9} = {r=2}{},
  cell{3}{1} = {r=4}{},
  cell{7}{1} = {r=4}{},
  cell{10}{3} = {c},
  cell{10}{4} = {c},
  cell{10}{5} = {c},
  cell{10}{6} = {c},
  cell{10}{7} = {c},
  cell{10}{8} = {c},
  cell{10}{9} = {c},
  vline{2-3} = {1-10}{},
  hline{1,7,11} = {-}{},
  hline{2} = {1-7}{},
  hline{3} = {1-9}{},
  hline{10} = {1-9}{},
}
Type    & Method          & Disc3D-QA             &                          &                      &                &                 & {ScanQA (val)\\(EM/EM-R)} & {SQA3D (test)\\(EM/EM-R)} \\
        &                 &  {Absolute \\Distance} & {Attribute\\Recognition} & {Relative\\Distance} & {Object\\Size} & {Object\\Count} &                           &                           \\
2D-MLLM & InternVL3-8B~\cite{zhu2025internvl3}    & 0.223                 & 0.649                    & 0.478                & 0.235          & 0.250           & --                        & --                        \\
        & InternVL3-14B   & 0.226                 & 0.708                    & 0.367                & 0.352          & 0.253           & --                        & --                        \\
        & InternVL3-38B   & 0.193                 & 0.711                    & 0.555                & 0.362          & 0.396           & --                        & --                        \\
        & InternVL3-78B   & 0.202                 & 0.702                    & 0.486                & 0.345          & 0.344           & --                        & --                        \\
3D MLLM & Ross3D~\cite{wang2025ross3d}          & 0.106                 & 0.672                    & 0.529                & 0.181          & 0.232           & \textbf{0.308 }/ --       & 0.630 / \textbf{0.657}    \\
        & Video-3DLLM~\cite{zheng2025video}     & 0.150                 & 0.638                    & 0.211                & 0.002          & 0.188           & 0.301 / --                & 0.586 /~--                \\
        & LLaVA-3D~\cite{zhu2024llava}        & 0.071                 & 0.590                    & 0.514                & 0.046          & \textbf{0.580}  & -- / 0.306                & -- / 0.601                \\
        & ~LLaVA-3D, ours & \textbf{0.299}        & \textbf{0.874}           & \textbf{0.730}       & \textbf{0.505} & 0.440           & 0.272 / 0.446             & 0.600 / 0.629             
\end{tblr}
\caption{Performances of different 2D MLLMs and 3D MLLMs on the Disc3D QA benchmark and other public 3D-QA benchmarks. We re-implement LLaVA-3D with InternVL3-8B as base model. Our Disc3D training corpus equips LLaVA-3D with consistent gains across multiple axes, as verified on the proposed Disc3D-QA suite and two public benchmarks. Although we deliberately excluded object-counting samples to mitigate dataset bias, the trained model still generalizes well to this task. 
}
\label{tab:sota_results}
\end{table*}
 
\section{Experiments}
\label{exp}

We evaluate on two canonical 3D scene-understanding tasks. (1) \textbf{3D visual grounding} localises the object described by a natural-language utterance.
Asking 3D MLLMs to regress a 3D bounding box in text space is unreliable.
We therefore benchmark proposal-matching methods~\cite{zheng2025video,wang2025ross3d} that recast the task as proposal classification: the model aligns predicted grounding tokens with 3D proposals extracted by an off-the-shelf detector~\cite{schult2022mask3d} at the feature level.
We report accuracy on the Disc3D grounding split with 2790 samples and analyse performance under varying conditions. (2) \textbf{3D question answering} uses a linear, objective metric that cleanly reflects different facets of model capability; consequently, our main experiments and ablations centre on QA.
Benchmarks include the public datasets~\cite{azuma2022scanqa,ma2022sqa3d} and the Disc3D multi-dimensional QA suite.
The Disc3D-QA test set contains 1\,000 questions per category across five dimensions; object counts are unavailable at training time, making this sub-task zero-shot for our model. 

\paragraph{Metrics.} For visual grounding, we report exact-match accuracy: we use ground-truth object boxes as proposals, thereby removing any confounding effect of localization error. For question-answering, we adopt top-1 exact-match (EM) and its relaxed variant (EM-R) for normal questions, and Mean Relative Accuracy~\cite{yang2025thinking} (MRA) for numerical questions such as object size and absolute distance.

\paragraph{Implementation Details.} We re-implement LLaVA-3D~\cite{zhu2024llava} on InternVL3~\cite{zhu2025internvl3} and fine-tune the MLP projector and LLM part. For our main experiments and ablations, leveraging its compact end-to-end design and competitive efficiency.
In addition to Disc3D, we combine 220\,K samples from public datasets to enlarge diversity and enable direct comparison with prior work (see \cref{tab:public_ds}). To balance cost and coverage we sample at most 32 frames per scene with max-coverage sampling~\cite{zheng2025video}; all images are centre-cropped to $336\times336$ pixels.

\begin{table}[t]
\centering
\small
\setlength{\tabcolsep}{1mm}
\renewcommand{\arraystretch}{1.05}
\begin{tblr}{
  colspec = {Q[l,1.2cm] Q[l,2.5cm] Q[c,1.5cm] Q[c,1.5cm]},
  cells = {m},
  cell{1}{1} = {r=2}{},
  cell{1}{2} = {r=2}{},
  cell{1}{3} = {c=2}{},
  cell{4}{1} = {r=2}{},
  hline{1,3-4,6} = {-}{},
}
Distractors & Objetc Reference          & Accuracy (\%) &        \\
            &                         & Video-3D-LLM~\cite{zheng2025video}  & Ross3D~\cite{wang2025ross3d} \\
No          & Normal        & 47.5          & 50.7   \\
Yes         & Discriminative & 27.3          & 28.7   \\
            & +~Anchor context        & 14.2          & 14.5   \\
\end{tblr}
\caption{Impact of category-similar distractors and referring-expression type on Disc3D grounding accuracy (proposal-matching methods). ``Discriminative'' referral encodes object size, spatial relation, and appearance (Comparative Disambiguation). ``+Anchor'' appends anchor-context to the expression (Spatially Anchoring).}
\label{tab:Disc3D_vg}
\end{table}

\begin{figure*}[t]
\centering
    \begin{subfigure}[b]{0.3\textwidth}
        \centering
        \includegraphics[width=\textwidth]{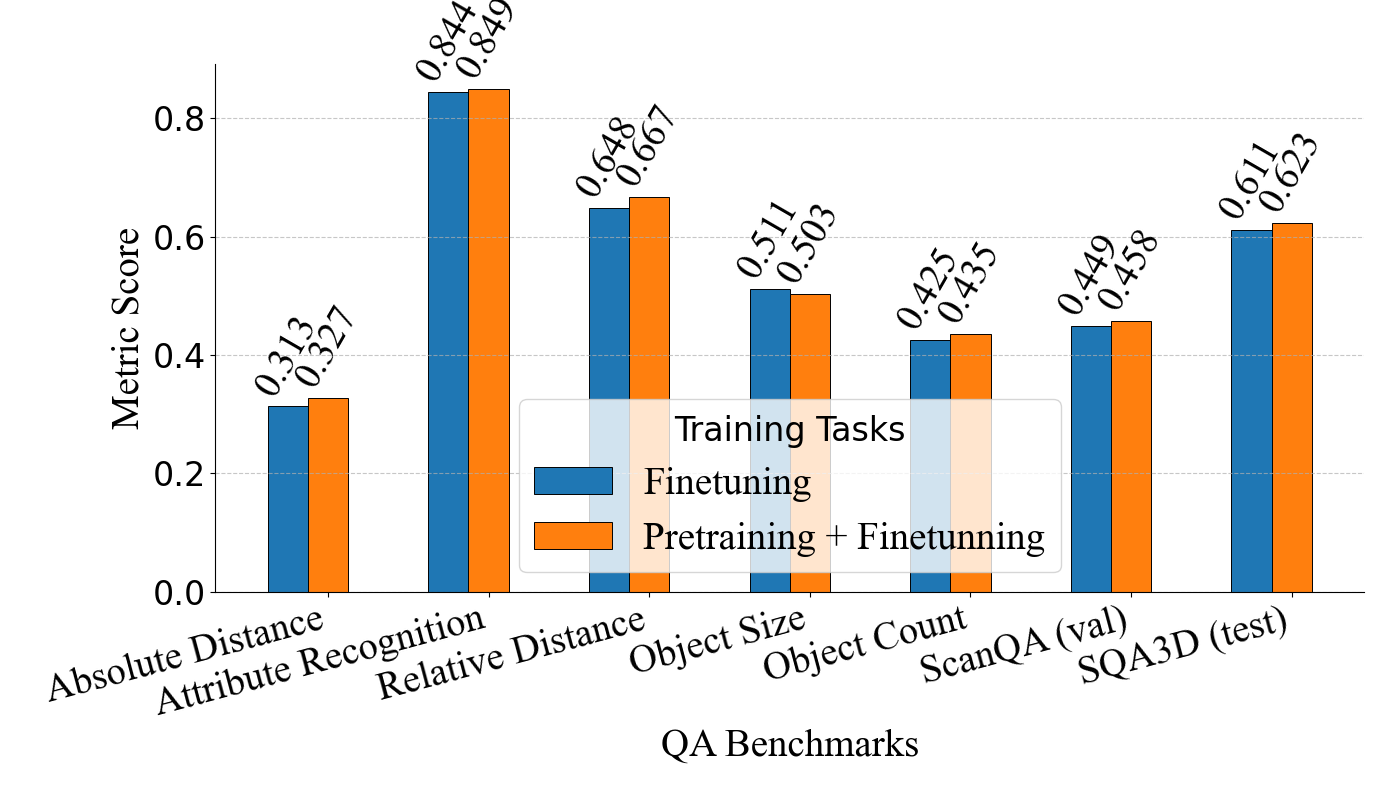}
        \caption{Comparison between different
training paradigms.}
        \label{fig:sub_a}
    \end{subfigure}
    \hfill
    \begin{subfigure}[b]{0.3\textwidth}
        \centering
        \includegraphics[width=\textwidth]{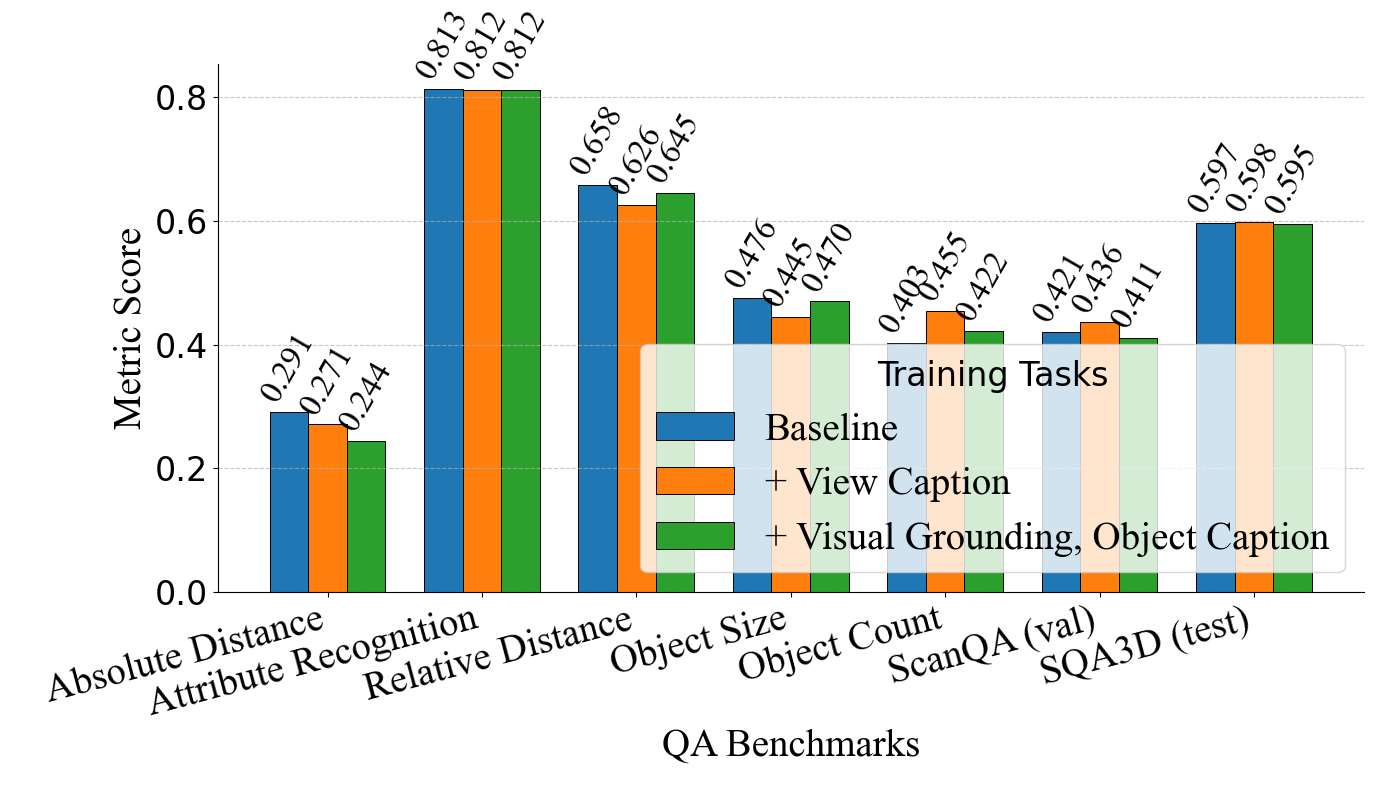}
        \caption{Comparison between different
training tasks.}
        \label{fig:sub_b}
    \end{subfigure}
    \hfill
    \begin{subfigure}[b]{0.3\textwidth}
        \centering
        \includegraphics[width=\textwidth]{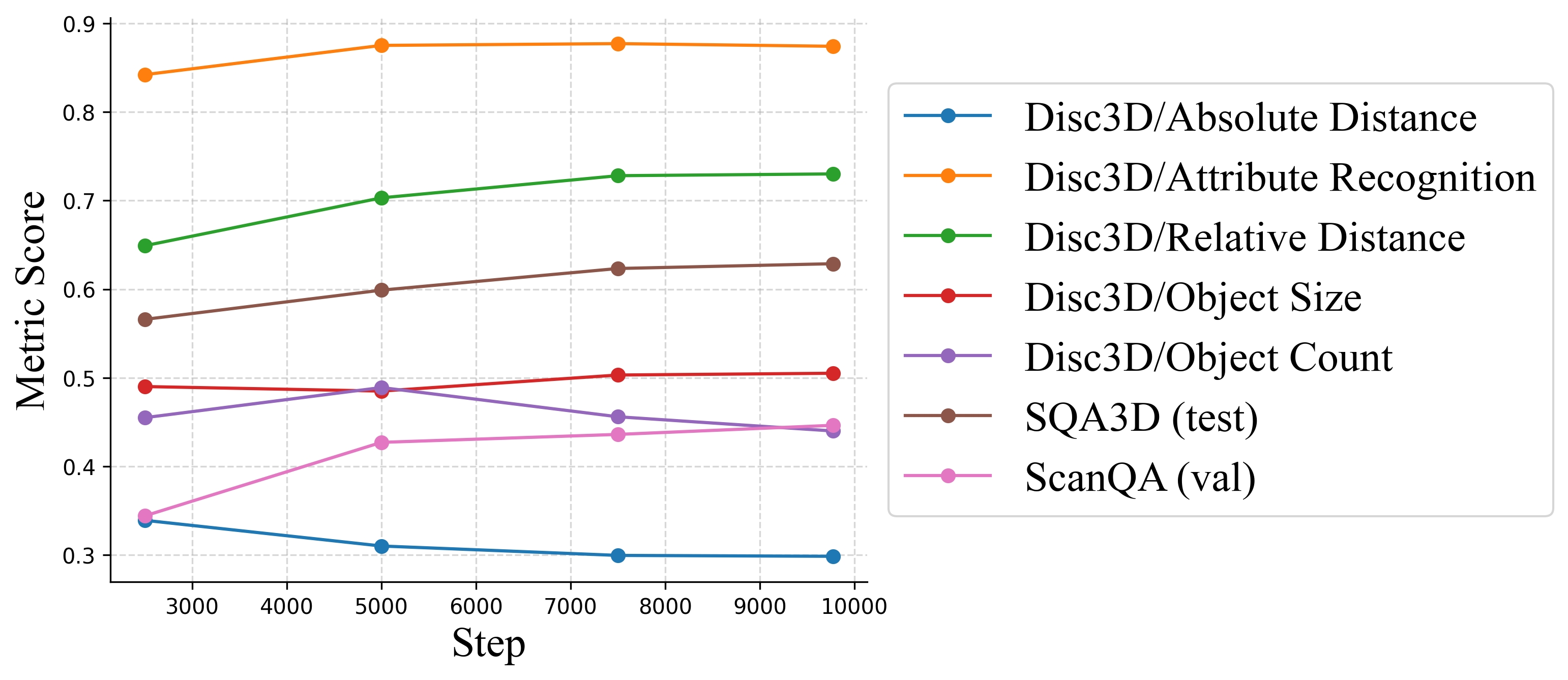}
        \caption{Evolution of performance on various
benchmarks as training progresses.}
        \label{fig:sub_c}
    \end{subfigure}

\caption{Results for ablation study. (a) Two-stage training yields consistent gains across most benchmarks; (b) Joint training with multi-task data improves the zero-shot performance on the object-counting task. (c) With the full Disc3D-QA dataset for fine-tuning, performance on most benchmarks rises with additional training steps, except for two numeric estimation tasks.
}
\label{fig:ablation_study}
\end{figure*}

\paragraph{3D MLLMs on the Disc3D visual-grounding task}
As \cref{tab:Disc3D_vg} shows, even with perfect localization, both Video-3D-LLM and Ross3D incur large accuracy drops as soon as distractors are present. Under the discriminative object-referral protocol, performance deteriorates further when the referring expression embeds richer spatial context generated by Spatially Anchoring.

\begin{table}[t]
\centering
\small
\setlength{\tabcolsep}{1mm}
\begin{tabular}{llll}
\toprule
\textbf{Dataset} & \textbf{Task} & \textbf{Size} & \textbf{Filtered} \\
\midrule
MMScan  & RegionCaption & 4724   & Yes \\
MMScan  & QA            & 115072 & Yes \\
SQA3D   & QA            & 79445  & No  \\
ScanQA  & QA            & 19046  & Yes \\
\bottomrule
\end{tabular}
\caption{Details of the 220K public dataset. We clean the dataset by applying keyword filters to remove samples exhibiting potential viewpoint ambiguity. SQA3D dataset is not cleaned, as its \textit{SituatedQA} task inherently encodes positional and viewpoint information within the context.}
\label{tab:public_ds}
\end{table}

\begin{table}[t]
\centering
\small
\setlength{\tabcolsep}{1mm}
\begin{tblr}{
  cell{2}{1} = {r=2}{},
  cell{4}{1} = {r=2}{},
  vline{2-3} = {1-5}{},
  hline{1-2,4,6} = {-}{},
}
Model Size & Training Data & {ScanQA \\(val)}   & {SQA3D \\(test)}   \\
2B         & Public 220K   & 0.312          & 0.569          \\
           & + Disc3D-QA      & \textbf{0.421} & \textbf{0.597} \\
8B         & Public 220K   & 0.405          & 0.600          \\
           & + Disc3D-QA      & \textbf{0.449} & \textbf{0.611} 
\end{tblr}
\caption{Impact of augmenting the training set with Disc3D under a $1:1$ mixture and equal iteration steps. The results are evaluated under the refined exact-match protocol. }
\label{tab:Disc3D_vs_public}
\end{table}

\paragraph{Effect of Disc3D as training data.}
Training on the open-source corpus alone (220K samples, \cref{tab:public_ds}) is compared with the same schedule augmented by Disc3D-QA.
To control for compute, the total number of iterations is fixed. \cref{tab:Disc3D_vs_public} shows that Disc3D brings consistent gains on both ScanQA and SQA3D.

\paragraph{Impact of Task-Mixing.}  \cref{fig:ablation_study}b reports the QA performance of LLaVA-3D-2B trained with different Disc3D task mixtures. Only the object-count task benefits; all other auxiliary tasks yield no systematic improvement.

\paragraph{Training Paradigm.} We compare single-stage fine-tuning with a two-stage scheme on LLaVA-3D-8B.
In the first stage, the MLP projector is warmed up on Disc3D-1M, constructed by balanced multi-task sampling from Disc3D.
In the second stage, projector and LLM are updated jointly.
\cref{fig:ablation_study}a shows that the two-stage approach outperforms the single-stage baseline.

\paragraph{Evaluation of 2D \& 3D MLLMs on Disc3D-QA.} We evaluate 2D and 3D MLLMs on each sub-task of Disc3D-QA. Larger 2D MLLMs improve on most questions, yet remain at chance for \emph{absolute} and \emph{relative distance} without 3D priors. For 3D MLLMs, current SOTA methods still under-perform in the zero-shot evaluation on Disc3D-QA. We observe that Video-3D-LLM underperform random guessing (50\%) on the multiple-choice task --- \emph{relative distance} --- because they fail to select an answer from the provided options (i.e. A/B). Fine-tuning on Disc3D lifts LLaVA-3D across all tasks.

\paragraph{Training Dynamics on the Full Disc3D-QA Corpus}
Two-stage instruction tuning of LLaVA-3D-8B is run on the full QA corpus; multi-benchmark scores are logged in the second stage. \cref{fig:ablation_study}c shows steady gains on Disc3D \emph{attribute recognition} and \emph{relative distance}, mirrored by ScanQA and SQA3D.
Object-count and the metric-estimation tasks plateau; the former lacks training examples, the latter suffers from sparse numeric supervision and coarse scene encoding (patch-centre coordinates + voxel pooling).

\section{Limitations and Conclusion}
\label{conclusion}
We propose \textit{Discriminative Object Referring} to simulate realistic object-referring across multiple aspects in dialogues. Our framework resolves ambiguities present in earlier datasets while raising task difficulty. Without manual annotation, we curate Disc3D dataset of million samples  spanning 25K scans.

Experiments demonstrate that training on our Disc3D significantly enhances QA performance on the benchmark across all metrics and yields consistent gains on public benchmarks. Current 3D MLLMs nevertheless remain weak in metric-estimation tasks, hindered by under-developed 3D vision encoders and more powerful 3D MLLM designs. Further progress demands co-evolution of architectures and data.

{
    \small
    \clearpage
    \bibliographystyle{ieeenat_fullname}
    \bibliography{main}
}
\clearpage
\setcounter{page}{1}
\maketitlesupplementary

\section{Details in Data Curation Pipeline}
\subsection{7-DOF Bounding Box Annotation}
As discussed in the main text, the 9-DOF bounding box is sensitive to point-cloud noise, which degrades subsequent distance and collision estimation and complicates model learning. As Algorithm~\ref{alg:7dof_obb}, we therefore refit 7-DOF bounding boxes using the provided point-cloud annotations. In Disc3D, the z-axis of every scan is already aligned with gravity, eliminating the need for explicit gravity-direction estimation. For certain source datasets (e.g., Structured3D, CA-1M) they omit per-object point-cloud labels, but directly provided 7-DOF 3D bounding boxes.
 
\begin{algorithm}[ht]
\caption{Compute 7-DoF Oriented Bounding Box}
\label{alg:7dof_obb}
\textbf{Input}: A object point cloud $P = \{(x_i, y_i, z_i)\}_{i=1}^N$. \\
\textbf{Output}: A 7-DoF bounding box: center $C$, size $S$, and yaw angle $\text{yaw}_L$.
\begin{algorithmic}[1]
\IF{$|P| < 3$}
    \STATE \textit{\# Fallback to Axis-Aligned Bounding Box for few points}
    \STATE $(C, S) \gets \text{AABB}(P)$
    \STATE $\text{yaw}_L \gets 0$
    \STATE \textbf{return} $(C, S, \text{yaw}_L)$
\ENDIF

\STATE \textit{\# Step 1: Project points onto the XY plane}
\STATE $P_{xy} \gets \{(x_i, y_i) \mid (x_i, y_i, z_i) \in P\}$

\STATE \textit{\# Step 2: Compute Minimum Area Bounding Rectangle (MABR)}
\STATE $(c_{xy}, (d_1, d_2), \theta_{cv}) \gets \text{MinAreaRect}(P_{xy})$

\STATE \textit{\# Step 3: Determine the yaw angle from the MABR}
\STATE $\text{yaw}_{d1} \gets -\text{to\_rad}(\theta_{cv})$
\STATE $\text{yaw}_{d2} \gets \text{yaw}_{d1} + \pi/2$

\STATE \textit{\# Assign L, W based on the longer side in the XY plane}
\IF{$d_1 \ge d_2$}
    \STATE $L \gets d_1$, $W \gets d_2$, $\text{yaw}_L \gets \text{normalize\_angle}(\text{yaw}_{d1})$
\ELSE
    \STATE $L \gets d_2$, $W \gets d_1$, $\text{yaw}_L \gets \text{normalize\_angle}(\text{yaw}_{d2})$
\ENDIF

\STATE \textit{\# Step 4: Compute center and height along the Z-axis}
\STATE $z_{\min} \gets \min(\{z_i \mid (x_i, y_i, z_i) \in P\})$
\STATE $z_{\max} \gets \max(\{z_i \mid (x_i, y_i, z_i) \in P\})$
\STATE $c_z \gets (z_{\min} + z_{\max}) / 2$
\STATE $H \gets z_{\max} - z_{\min}$

\STATE \textit{\# Assemble the final 7-DoF box parameters}
\STATE $C \gets (c_{xy}[0], c_{xy}[1], c_z)$
\STATE $S \gets (L, W, H)$

\STATE \textbf{return} $(C, S, \text{yaw}_L)$
\end{algorithmic}
\end{algorithm}

\subsection{Label Graph Construction}
To construct the label graph, we adopt DeepSeek-V3~\cite{liu2024deepseek} as the oracle for deciding subsumption  between object labels. The query template is:

\textit{You are an expert in indoor/outdoor object categorization. Answer YES or NO: the category ``\{label1\}'' is a subclass of ``\{label2\}''.}

\subsection{Details of Source Dataset Processing} 
\begin{figure}[ht]
\centering
\includegraphics[width=0.47\textwidth]{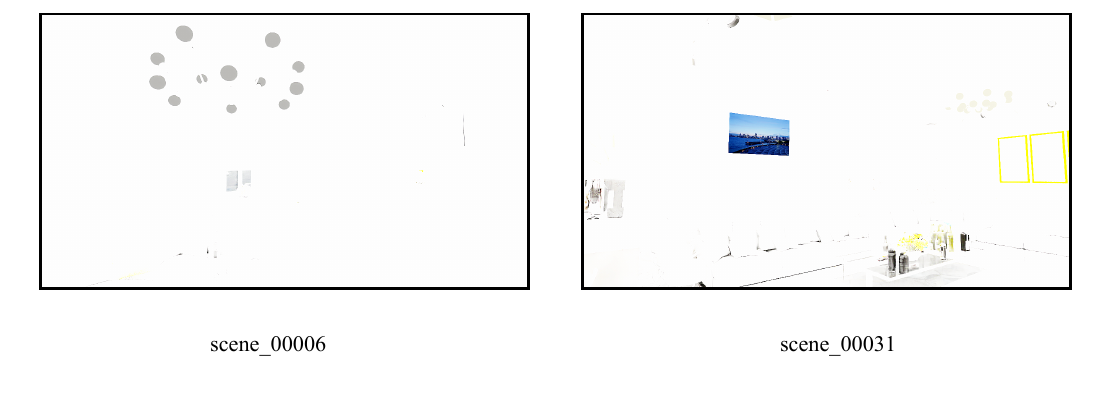}
\caption{Samples of overexposed images from the Structured3D dataset.}
\label{fig:over_exp_cases}
\end{figure}

Beyond the issues discussed in the main text, several datasets require additional handling:

\begin{enumerate}
    \item \textbf{Class-agnostic annotations in CA-1M.} CA-1M~\cite{lazarow2025cubify} provides object annotation without semantic labels. During the Object-Caption stage, we prompt a 2D MLLM with the object views and ask it to predict the class label, following the NYU-40 taxonomy.
    \item \textbf{Invalid scenes in Structured3D.} We filter out the scenes listed in the official error list. We further discard any image whose mean intensity exceeds a brightness threshold, thereby removing over-exposed views. Specifically, for each image, we compute the fraction of uint8 pixels whose gray-level intensity exceeds 245. If this fraction surpasses 90\%, the image is labeled as over-exposed. Following this criterion, we discard 143 over-exposed scenes from Structured3D~\cite{zheng2020structured3d}; representative examples are shown in Figure~\ref{fig:over_exp_cases}.

    \item \textbf{Missing segmentation labels in CA-1M.} Because CA-1M~\cite{lazarow2025cubify} lacks 2D and 3D segmentation, we estimate object visibility masks used in the caption annotation procedure heuristically. Inspired by OpenYolo3D~\cite{boudjoghra2024open}, we combine depth values with the provided 2D bounding boxes to construct a per-frame label map. For each image, we compute the average depth inside every bounding box, sort the boxes from far to near, and assign object IDs in this order. Consequently, regions of distant objects are covered by nearer ones.
\end{enumerate}

\subsection{MLLMs and LLMs for Annotation}

In most tasks, we rely on proprietary MLLMs and LLMs for annotation, text rewriting, and other related operations. When the output must follow a specific structure, we include the desired format in the prompt and validate the returned result. The prompts for the key steps are provided below for reference (note that the content enclosed in double square brackets serves as a placeholder, awaiting the insertion of external variables at runtime.
):
\begin{itemize}
    \item Prompt for scene annotation: \textit{Given the multiple images of a scene, describe the scene. Note that you must obey the following requrirements: 1. Avoid using expressions like 'the image' in the description;2. Avoid using relative direction words/phrases, such as left/right/front/back;3. Do not give the specific number of any objects in the image and just use 'multiple';4. Be concise and accurate and output no more than 300 words.}
    \item Prompt for frame annotation: \textit{Describe the scene shown in the image. Note that you must obey the following requrirements: 1. Avoid using expressions like 'the image' in the description; 2. Avoid using relative direction words/phrases, such as left/right/front/back; 3. Do not give the specific number of any objects in the image and just use 'multiple';4. Be concise and accurate and output no more than 100 words.} 
    \item Prompt for object annotation: \textit{Suppose you are an image annotation assistant, and you will judge whether the "[[ObjectLabel]]" can be seen in the bounding box of given images and describe it. Follow the following rules in one step:
    1. Firstly, judge that whether the "[[ObjectLabel]]" can be seen in the bounding box of given images and return YES/NO.
    2. If the "[[ObjectLabel]]" can be seen clearly, summarize a precise object description for "[[ObjectLabel]]".
    Note the following requirements:
    1. Avoid using words such as "image/scene" in the answer.
    2. Do not include other objects in the answer.
    3. Avoid using non-sense words, such as 'unclear', 'unknown', 'unidentified', 'uncertain', 'unpredictable', 'doubtful', 'undetermined', 'unable'.}
    \item Prompt for relation judgment and re-description of scene graph: \textit{Carefully observe the two objects enclosed by the bounding boxes in the image. Is the [[ObjectA]] [[Relation]] [[ObjectB]] in this picture? Return Yes/No, and if NO, briefly describe the spatial relationship between two objects, with the following requirements: 1. Do not use any descriptions that rely on the current camera perspective, such as 'in front of', 'behind', 'left' or 'right'; 2. Use the relations in [[Predefined Relations]].}
    \item Prompt for extracting the corrected the relation: \textit{Given the relationship description between objectA '[[ObjectA]]' and objectB '[[ObjectB]]': [[the corrected description of object relations]]
    Extract the preposition/phrase indicating their relationship, with the following requirements:
    1. Return The relationship preposition/phrase only and do not add any other words.
    2. The relationship preposition/phrase should satisfy the format: objectA is/are [relationship preposition/phrase] objectB.
    3. Avoid using words that depend on the current viewing perspective, such as 'front', 'back', 'behind', 'left' or 'right'}
\end{itemize}
We refine object-relation extraction by leveraging proprietary reasoning-style LLMs, thereby enhancing accuracy.

\begin{algorithm}[t]
\caption{Calculate Signed Angle Between Two 2D Vectors}
\label{alg:cal_angle}
\textbf{Input} Two 2D vectors $\mathbf{v}_1 = (x_1, y_1)$ and $\mathbf{v}_2 = (x_2, y_2)$.
\textbf{Output} The signed angle $\theta$ from $\mathbf{v}_1$ to $\mathbf{v}_2$ in radians, where $\theta \in [-\pi, \pi]$.
\begin{algorithmic}[1]
\STATE $\text{dot\_product} \gets x_1 x_2 + y_1 y_2$ 
\STATE $\text{cross\_product} \gets x_1 y_2 - y_1 x_2$
\STATE $\theta \gets \text{atan2}(\text{cross\_product}, \text{dot\_product})$
\STATE \textbf{Return} $\theta$
\end{algorithmic}
\end{algorithm}

\subsection{Comparative Disambiguation}
\begin{figure*}[!htp]
\centering
\includegraphics[width=0.8\textwidth]{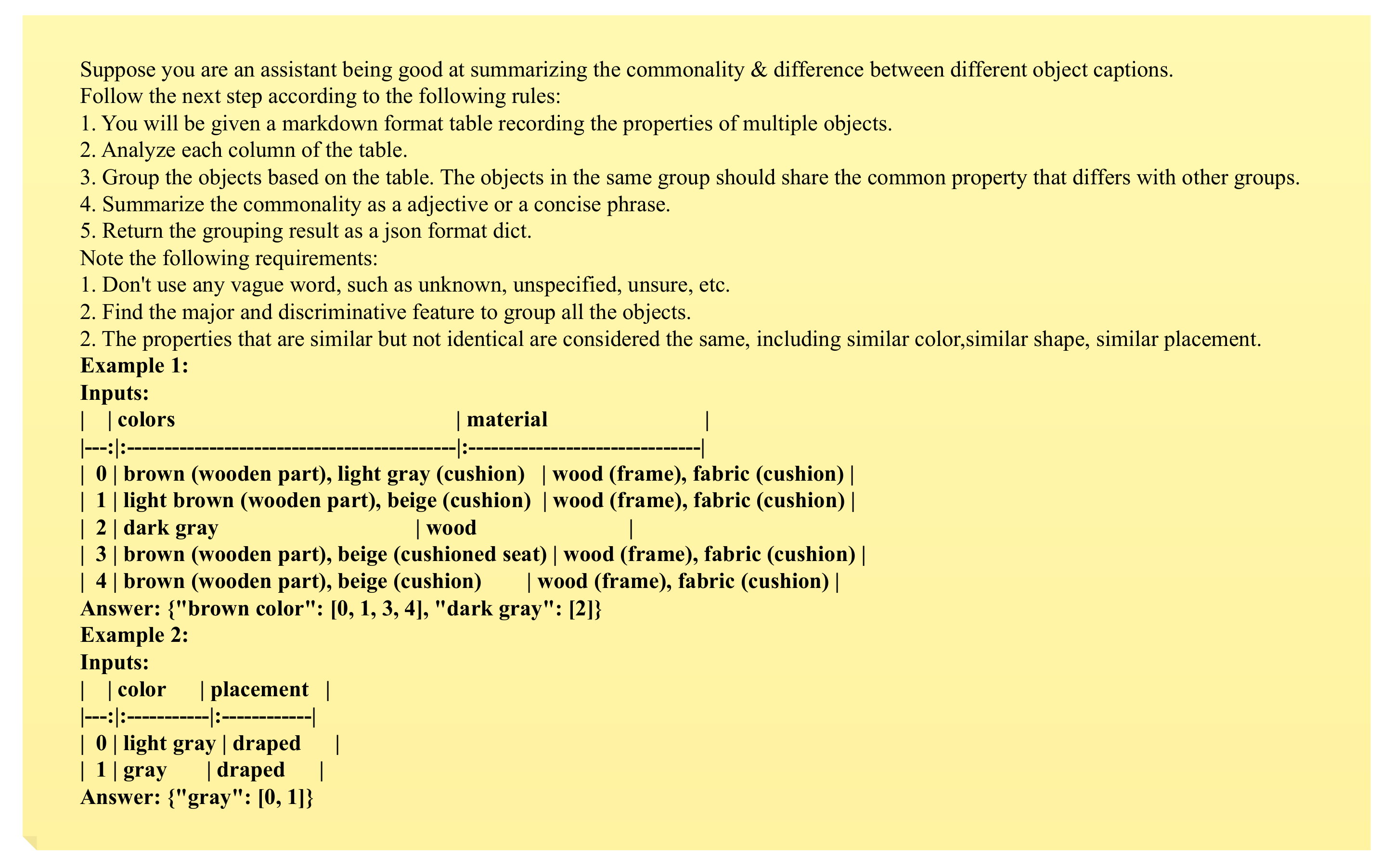}
\caption{Prompt for grouping distractors by appearance in the procedure of Comparative Disambiguation.}
\label{fig:group_by_app}
\end{figure*}

For appearance analysis, we leveraged an LLM assistant to extract and compare the color, shape, material, and condition attributes of objects within each distractor set (when available).  
As Figure~\ref{fig:group_by_app} shows, these attributes were compiled into markdown tables, and a small set of annotated cases was provided to steer the model toward accurate predictions via in-context learning.

Regarding object size, we assign the labels \texttt{largest} and \texttt{smallest} to the extreme instances (after accounting for a noise buffer), and \texttt{not-largest} and \texttt{not-smallest} to all others.

For object relations, within each resulting group, we identify the distinctive relations that distinguish a given object from the remaining members. To suppress noise, we assume that object-centric subgraphs are identical for any object pair whose members lie within 50 cm of each other. To keep the generated description concise, only one distinctive relation is retained.

\begin{figure}[!h]
\centering
\includegraphics[width=0.4\textwidth]{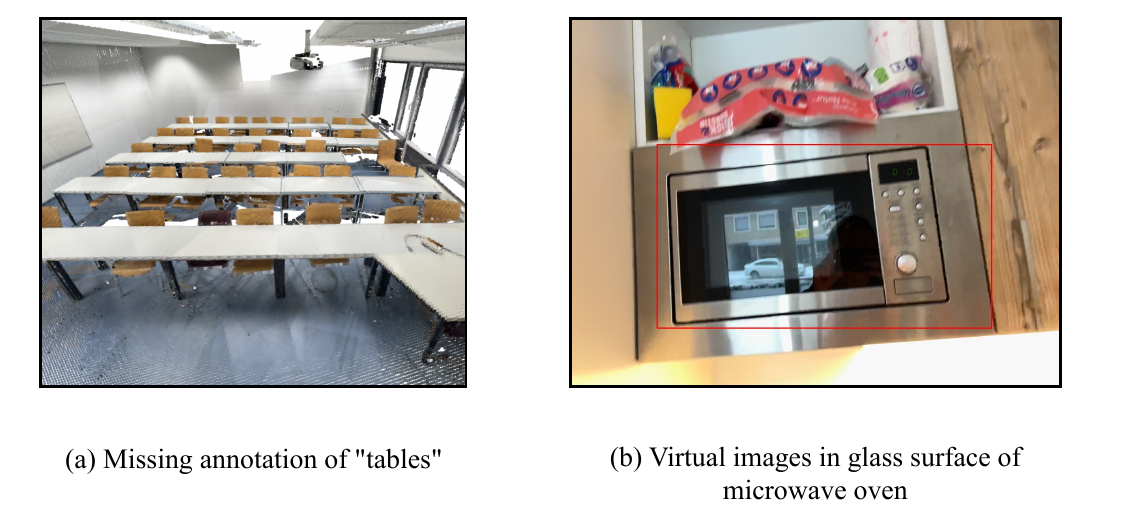}
\caption{Examples of remaining issues in 3D scene dataset.}
\label{fig:remaining_issues}
\end{figure}

\begin{figure}[!h]
\centering
\includegraphics[width=0.45\textwidth]{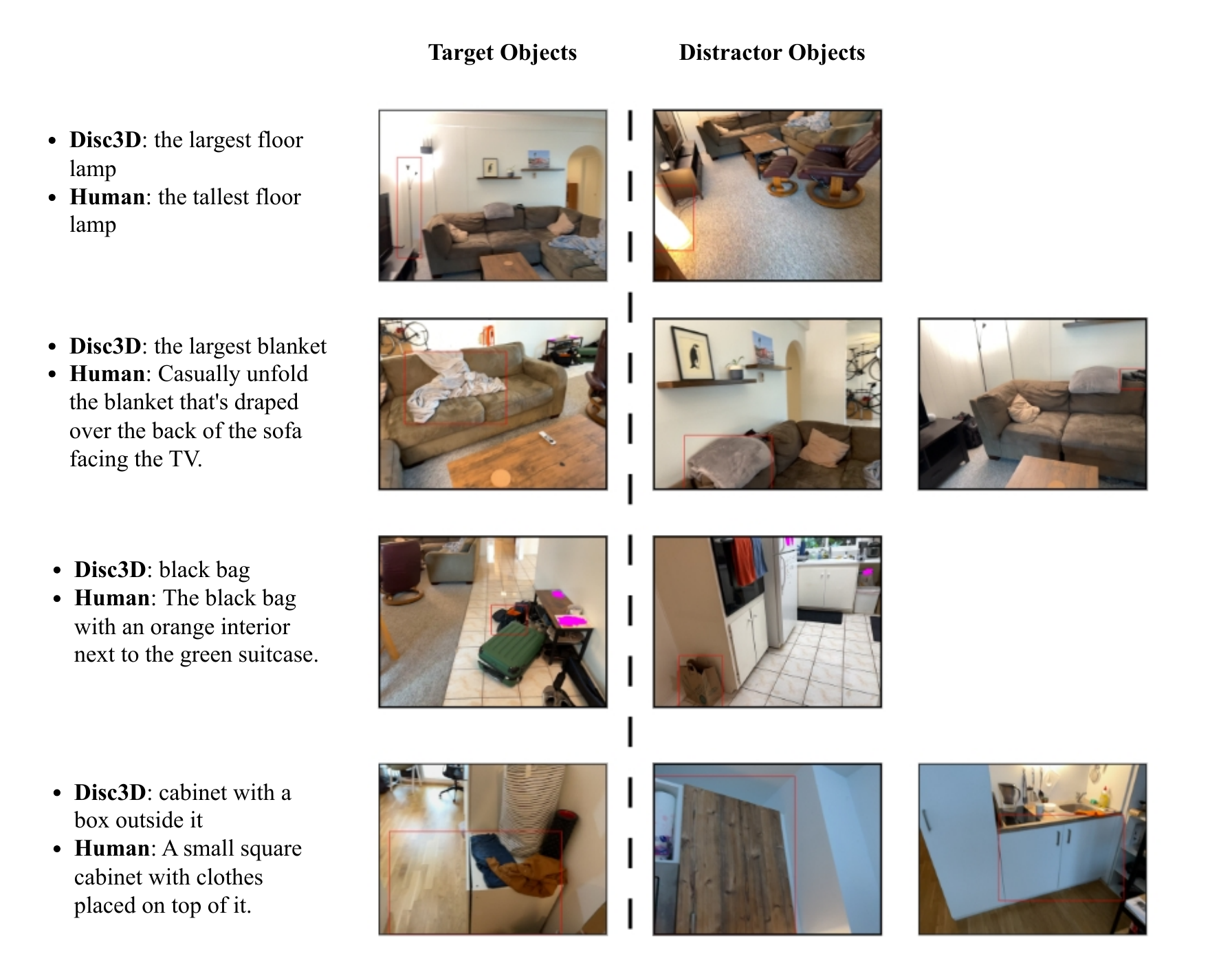}
\caption{Comparison of object descriptions before and after manual modification.}
\label{fig:human_check}
\end{figure}


\subsection{Object Distance Estimation}
To approximate the distance between two objects, we adopt their 7-DOF bounding boxes as coarse spatial proxies.  
The procedure is as follows:

\begin{enumerate}
  \item Detect overlap by applying the Separating Axis Theorem (SAT) to the two boxes. If an overlap is detected, the inter-object distance is set to zero.
  \item Otherwise, sample 16 uniformly distributed points on each face of both boxes and compute all pairwise point-to-point distances; the minimum among these values is taken as the estimated distance between the objects.
\end{enumerate}

\subsection{Spatially Anchoring}
For Spatially Anchoring, we generate discriminative object descriptions according to the following logic:
\begin{enumerate}
  \item \textbf{Anchor‐object sampling:} We sample from the non‐distractors of current distractor group, which must lie at least 50\,cm from the anchor and compute all pairwise distances between the anchor object and the target. If an object is the furthest or the nearest to the anchor, we assign it the corresponding template‐based description. To account for estimation errors in 3D bounding‐box distances, we introduce a dynamic noise buffer equal to the maximum dimension among all orientation‐aligned bounding boxes in the distractor group.
  \item \textbf{Anchor‐sight sampling:} For objects defining the start and end of the anchor sight, we enforce a minimum separation of 50\,cm. To eliminate the effect of height differences, we consider only the XY plane when determining the relative direction between an object and the sightline. For each distractor group, if an object lies at the extreme right or extreme left of the sightline (within a $10^\circ$ buffer), we assign the corresponding descriptor. The computation of the object’s relative angle to the sightline is illustrated in Algorithm~\ref{alg:cal_angle}.
\end{enumerate}

\begin{figure*}[!htp]
\centering
\includegraphics[width=0.95\textwidth]{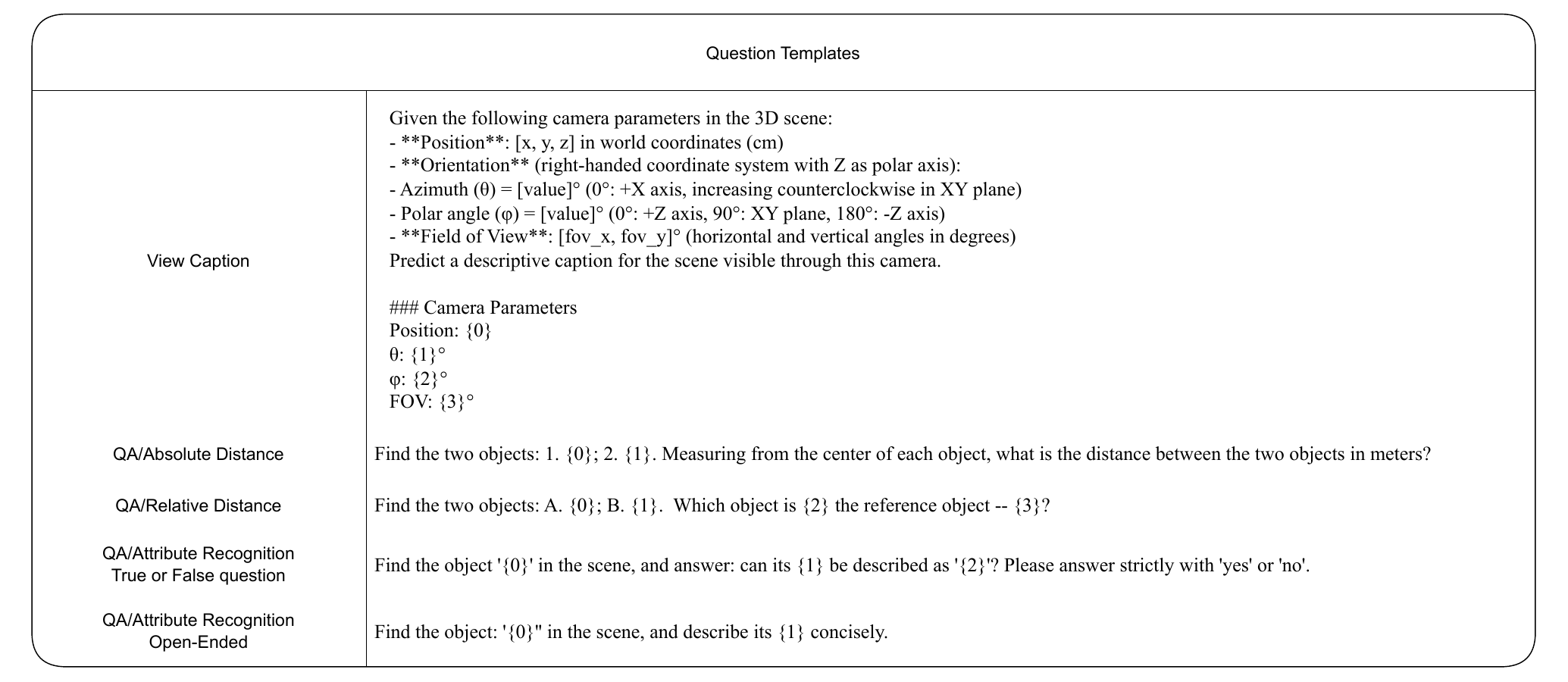}
\caption{Question templates for multi-task data generation.}
\label{fig:question_templates}
\end{figure*}

\begin{figure*}[!htp]
\centering
\includegraphics[width=0.95\textwidth]{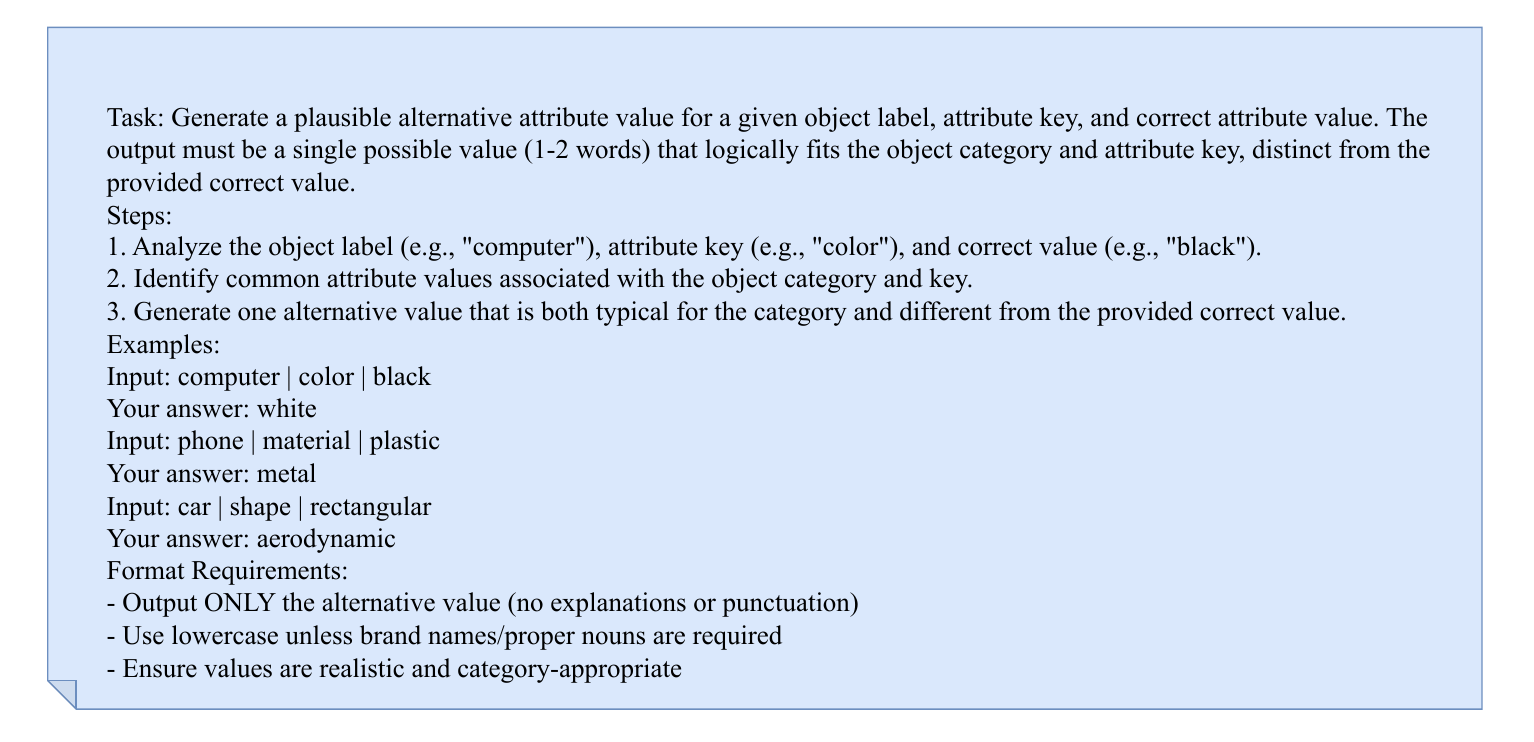}
\caption{The prompt for guiding LLM to generate the distracting object attribute in QA.}
\label{fig:llm_confuse}
\end{figure*}

\subsection{Details for Attribute Recognition QA}
As shown in Table~\ref{tab:llm_confuse}, for true/false questions in attribute‐recognition QA, directly sampling incorrect answers from attributes of other objects in the same category cannot eliminate the influence of semantically similar attributes expressed in different words. Therefore, we employ a large language model (LLM) to generate intentionally misleading wrong answers. As illustrated in Figure~\ref{fig:llm_confuse}, we supply the model with the object label, attribute field, and attribute value, guiding the LLM to produce plausible yet non-identical attribute values for that object category. This strategy helps mitigate the shortcut risk in QA tasks --- namely, the tendency of the model to infer answers solely from the object type mentioned in the question.

\begin{table*}[t]
\centering
\begin{tblr}{
  cell{1}{1} = {r=2}{},
  cell{1}{2} = {c=3}{},
  cell{1}{5} = {c=3}{},
  vline{2-3} = {1}{},
  vline{5} = {1-7}{},
  vline{2,5} = {1-7}{},
  hline{1,3,8} = {-}{},
  hline{2} = {2-7}{},
}
Catergory & Attributes    &          &        & Distracting Attributes &          &        \\
          & color         & material & shape  & color                  & material & shape  \\
curtain   & white         & --       & --     & blue                   & --       & --     \\
door      & brown         & wooden   & --     & white                  & metal    & --     \\
carpet    & black         & --       & --     & gray                   & --       & --     \\
tv        & black         & --       & flat   & silver                 & --       & curved \\
pillow    & light-colored & --       & square & blue                   & --       & round  
\end{tblr}
\caption{Comparison between the object attributes and the distracting attributes generated by the LLM}
\label{tab:llm_confuse}
\end{table*}

\subsection{Question Templates}
For Scene Caption \& Visual Grounding, the question prompts for the two tasks are generated by LLM for summarizing the room descriptions. Other question templates can be found in Figure~\ref{fig:question_templates}. 
For attribute-recognition tasks, the training set comprises both true/false questions and open-ended items, whereas the test set is restricted to true/false questions to enable a more precise evaluation.

\begin{figure*}[!htp]
\centering
\includegraphics[width=0.95\textwidth]{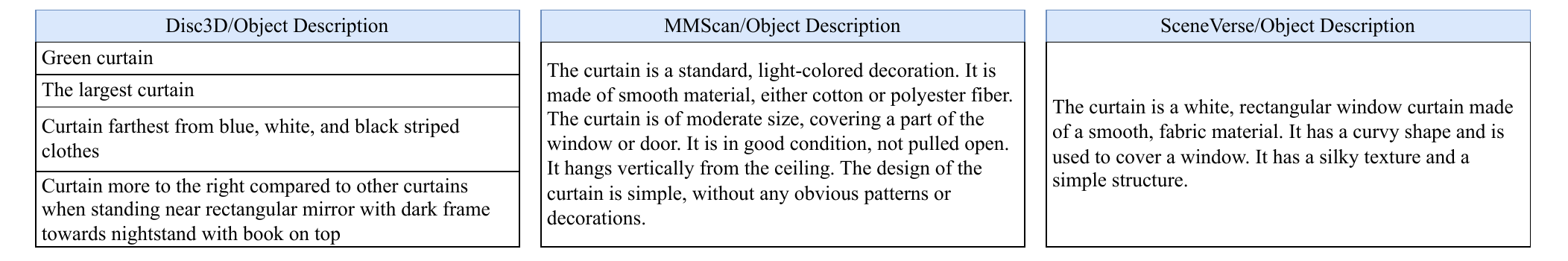}
\caption{A comparison of object descriptions across different datasets (using "curtain" as an example).}
\label{fig:cmp_diff_ds_obj_descrip}
\end{figure*}

\subsection{Comparison of Object Descriptions}
As shown in Figure~\ref{fig:cmp_diff_ds_obj_descrip}, we provide a comparison of object descriptions.
MMScan~\cite{lyu2024mmscan} and SceneVerse~\cite{jia2024sceneverse} enumerate almost every observable attribute --- appearance, location, geometry, functionality, and ancillary details, yielding exhaustive yet lengthy descriptions.
Such verbosity, while informative, dilutes the discriminative cues required to unambiguously identify an object.
In contrast, our Disc3D pipeline distills a concise, multi-view signature that foregrounds the distinguishing characteristics of each object, thereby offering a more effective dataset for assessing and guiding the spatial understanding capability of 3D MLLMs.

\subsection{Remaining Limitations}
We addressed a range of defects in the source dataset, including subsumption relations in semantic labels, abnormal camera poses, blurred images, and point-cloud noise. Despite these remedies, two issues remain unresolved.
\begin{enumerate}
\item \textbf{Missing object annotations.} Dense or poorly delineated objects are often skipped, leading to incomplete labels (see Figure~\ref{fig:remaining_issues}(a)).
\item \textbf{Reflection and refraction.} Glass surfaces create virtual images that MLLMs incorporate into view-dependent descriptions, thereby distorting the 3-D scene understanding (see Figure~\ref{fig:remaining_issues}(b)).
\end{enumerate}

\section{Disc3D benchmark}
We provide Disc3D-QA, the benchmark comprising 5K annotated examples evaluated in the main paper, in the supplementary material.

\subsection{Human Annotation for object referrals in Disc3D }

As shown in Figure~\ref{fig:human_check}, we present representative cases before and after human quality inspection.
Factual inaccuracies in descriptions generated by the Comparative Disambiguation module of our Disc3D pipeline (e.g., the fourth row in Figure~\ref{fig:human_check}) occur far less frequently than the edits introduced by annotators.
A substantial portion of revisions target stylistic or linguistic refinement rather than correctness, with a notable share focusing on size descriptors (rows one and two in (e.g., the fourth row in  Figure~\ref{fig:human_check})).
Simple tokens such as \texttt{largest} or \texttt{smallest} often fail to capture the distinctive size characteristics of objects with irregular shapes (e.g., floor lamps) or soft, deformable items (e.g., blankets).


\end{document}